\documentclass{article}

\usepackage[preprint, nonatbib]{nips_2018}

\usepackage{float}

\usepackage[square,numbers]{natbib}
\usepackage[utf8]{inputenc} % allow utf-8 input
\usepackage[T1]{fontenc}    % use 8-bit T1 fonts
\usepackage{hyperref}       % hyperlinks
\usepackage{url}            % simple URL typesetting
\usepackage{booktabs}       % professional-quality tables
\usepackage{amsfonts}       % blackboard math symbols
\usepackage{nicefrac}       % compact symbols for 1/2, etc.
\usepackage{microtype}      % microtypography
\usepackage{graphicx}
\usepackage{tabularx}
\usepackage{hyperref}

\usepackage{amsmath}
\usepackage{algorithm}  
\usepackage{algpseudocode} 
\usepackage{subfigure}
\usepackage{multirow}
\usepackage{makecell}

\title{Self-supervised Learning for Label Sparsity in Computational Drug Repositioning}

\author{
	Xinxing Yang \\
	
	Ningbo Artificial Intelligence Institute, Shanghai Jiao Tong University\\
	Department of Automation, Shanghai Jiao Tong University\\
	\texttt{yangxinxing@sjtu.edu.cn} \\
	
	\And
	Genke Yang \\
	Ningbo Artificial Intelligence Institute, Shanghai Jiao Tong University\\
	Department of Automation, Shanghai Jiao Tong University\\
	\texttt{gkyang@sjtu.edu.cn} \\
	
	\And
	Jian Chu \\
	Ningbo Artificial Intelligence Institute, Shanghai Jiao Tong University\\
	Department of Automation, Shanghai Jiao Tong University\\
	\texttt{chujian@sjtu.edu.cn} \\
	
}

\begin{document}
\maketitle

\begin{abstract}
	
The computational drug repositioning aims to discover new uses for marketed drugs, which can accelerate the drug development process and play an important role in the existing drug discovery system. However, the number of validated drug-disease associations is scarce compared to the number of drugs and diseases in the real world. Too few labeled samples will make the classification model unable to learn effective latent factors of drugs, resulting in poor generalization performance. In this work, we propose a multi-task self-supervised learning framework for computational drug repositioning. The framework tackles label sparsity by learning a better drug representation. Specifically, we take the drug-disease association prediction problem as the main task, and the auxiliary task is to use data augmentation strategies and contrast learning to mine the internal relationships of the original drug features, so as to automatically learn a better drug representation without supervised labels. And through joint training, it is ensured that the auxiliary task can improve the prediction accuracy of the main task. More precisely, the auxiliary task improves drug representation and serving as additional regularization to improve generalization. Furthermore, we design a multi-input decoding network to improve the reconstruction ability of the autoencoder model. We evaluate our model using three real-world datasets. The experimental results demonstrate the effectiveness of the multi-task self-supervised learning framework, and its predictive ability is superior to the state-of-the-art model.
	
\end{abstract}

%\keywords{Computational Drug Repositioning \and Collaborative Filtering \and Outer Product \and Positive-Unlabeled Learning \and Curative Effect Prediction }

\section{Introduction}

Traditional drug discovery has a long R\&D cycle, high investment cost, high risk and low success rate \cite{1,2}. It takes roughly 10-15 years and \$0.8-\$1 billion to bring a new drug to market from development \cite{3,4}. The difficulty of new drug development is that about 90\% of new drugs do not pass Phase I clinical trials because their new chemical structures lead to unpredictable side effects during actual use. Therefore, there is a need for a new technology that can accelerate the drug development process and ensure that drug discovery has a high success rate and low risk \cite{5,6}. \par

Computational drug repositioning aims to uncover new uses for marketed drugs based on known drug-disease associations \cite{7,8}. The logic behind it is that the small molecule drugs currently on the market have multi-target properties, which means that they can inhibit or activate unknown targets, thereby producing therapeutic effects on unknown diseases. Computational drug repositioning uncovers potential therapeutic patterns of drugs and diseases through computational models and the large number of validated drug-disease associations. Based on these patterns, new therapeutic uses of the target drug can be inferred \cite{9,10}. \par

The popular computational drug repositioning models can be divided into two categories, namely graph-based model \cite{11,12,13,14} and matrix factorization-based model \cite{15,16,17}. The first step of the graph-based model is to construct a heterogeneous network based on multi-source information of drugs and diseases, and then use algorithms such as random walks to mine potential drug-disease associations on the above heterogeneous network. Wang et al. \cite{n1} integrated information from multiple sources, including omics information of diseases, drugs and targets, to construct a heterogeneous graph. A random walk algorithm is then performed on the entire heterogeneous graph to compute potential drug-disease associations. Luo et al. \cite{n2} constructed a heterogeneous graph based on drug similarity information, disease similarity information and drug-disease associations. The Bi-Random walk algorithm was then used to walk through the heterogeneous graph to predict new drug-disease associations. Zeng et al. \cite{n3} proposed a network-based deep learning method, deepDR. The model integrates multiple drug and disease-related networks for mining new uses of drugs. These networks are the drug-disease association network, the drug side effect network, the drug-target network, and the multiple drug interaction network. The deepDR model is used to predict the probability of a therapeutic relationship between drugs and diseases by learning higher-order features in these networks through multiple autoencoder models. Previous graph-based models assume that neighbors are independent of each other in heterogeneous graphs, which leads to the loss of local structural information. Therefore, Meng et al. \cite{n4} performed graph convolution operations on drug-disease association networks, drug-drug similarity networks and disease-disease similarity networks to learn a unified representation of drugs and diseases. Then the drug and disease representations are input into a multi-layer fully-connected network with network regularization elements to obtain drug-disease association probabilities. \par

The idea of the matrix factorization-based model is to represent the latent factor of drugs and diseases using vectors in embedding space. The probability of a therapeutic association between a drug and a disease is subsequently calculated by the similarity function. Zhang et al. \cite{m1} proposed a computational model based on Bayesian inductive matrix for mining new uses of marketed drugs. Unlike previous work that used data from a single source to calculate drug and disease similarity, DRIMC used multiple data sources. In addition, the features of the drugs (diseases) are considered with information about their respective neighbors. These features were then combined with the induction matrix to obtain the predicted values. Yan et al. \cite{m2} proposed a multi-view matrix factorization model for predicting novel drug-disease associations. The model mines the combined drug/disease similarity matrix from multiple sources of drug/disease structural information. And this similarity matrix is combined with the drug-disease association matrix to derive the treatment probability between the target drug and the target disease through regularization techniques and multi-view learning. Yan et al. \cite{m3} argued that the linear representation of the matrix factorization model has limitations and cannot capture the complex relationships between drugs and diseases. Therefore, they proposed the ASMF model, which utilizes an attention mechanism instead of the inner product operation, enabling the model to take into account the unique weights of each feature. And they use the improved autoencoder model to extract more effective features of drugs and diseases. The matrix factorization model uses the inner product operation and vectors in embedding space to represent the relationship between drugs and diseases. Yang et al. \cite{m4} demonstrated that this approach does not represent the relationship between drugs and diseases correctly. Therefore, they used a modified Euclidean distance and point space to represent the drug-disease relationship and demonstrated that this approach is superior to the inner product operation and vector space on several real data sets. \par

The focus of this paper is on the two-tower model (matrix factorization-based model) in computational drug repositioning, where the main idea is to use two different sets of neural networks to learn the latent factor of the drug and disease. The degree of compatibility of the two latent factors is subsequently calculated using a similarity function (inner product) as the probability that the drug can treat the disease. This class of models is described in detail in section 2.1. \par

\begin{table}[h]
	\caption{The statistics of three real-world datasets}
	\centering
	\setlength{\tabcolsep}{2.5mm}
	\begin{tabular}{@{}ccccc@{}}
		\toprule
		Datasets  & Drugs & Diseases & Validated Associations & Sparsity \\ \midrule
		Gottlieb  & 593   & 313      & 1933                   & 98.95\%   \\
		Cdataset  & 663   & 409      & 2532                   & 99.06\%   \\
		DNdataset & 1490  & 4516     & 1008                   & 99.98\%   \\ \bottomrule
	\end{tabular}
\end{table}

\begin{figure}[h]
	\centering
	\includegraphics[scale=0.87]{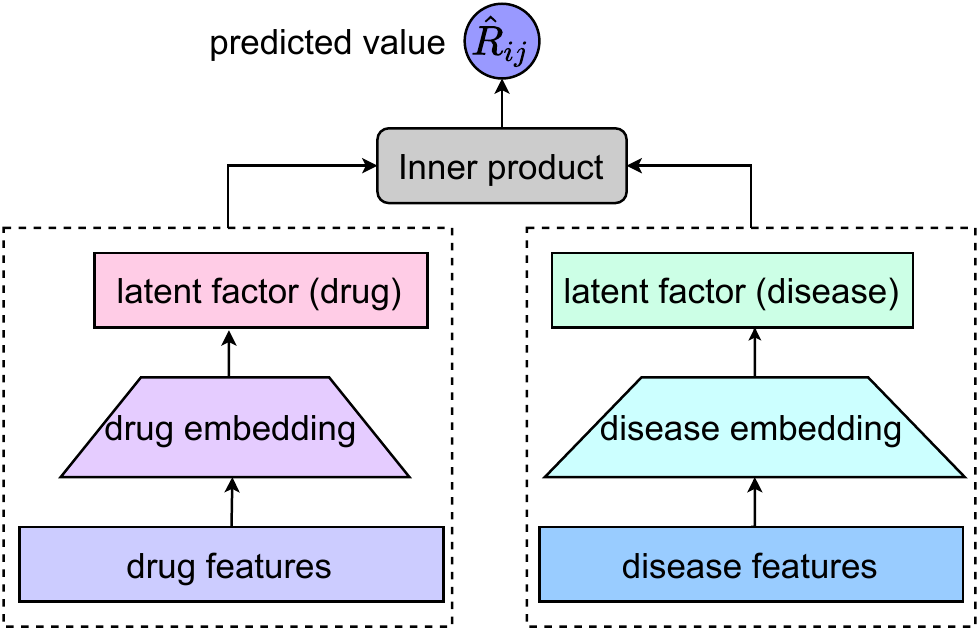}
	\caption{The workflow of two-tower model.}
	\label{}
\end{figure} 

The key to the usefulness of this model is to find the appropriate latent factor vector to represent the drug and disease. However, as the number of layers of the neural network increases, the number of parameters to be learned also increases exponentially. It is worth noting that the loss function used to train these models is usually formulated as a supervised learning problem. These supervised signals are sourced from validated drug-disease associations. The sparsity of the dataset used in this paper is shown in table 1, which reveals that validated drug-disease associations are very scarce compared to the larger number of drugs and diseases in the real world. This also means that it is difficult to train the large number of parameters in the two-tower model with these few labeled samples. \par

Inspired by the successful application of self-supervised learning in computer vision \cite{ssl}, we believe that self-supervised learning can provide a different perspective to enhance the representation of drugs via unvalidated data. Therefore, in this work, we propose a multi-task self-supervised learning framework (SSLDR) for tackle the label sparsity problem in computational drug repositioning. Specifically, we take the drug-disease association prediction problem as the main task, and the auxiliary task (self-supervised learning) is to use data augmentation strategies and contrast learning to mine the internal relationships of the original drug features, so as to automatically learn a better drug representation without supervised labels. The auxiliary task of the SSLDR framework can be divided into three steps. The first step is to use the similarity information to filter out the negative neighbors of the target drug. The second step is representation learning for the target drug and negative neighbors using two different data augmentation strategies. The third step is to map the above four drug representations into the embedding space so that the latent factors belonging to the same drug are close to each other in the embedding space, and those not belonging to the same drug are far away. In this way, a better latent factor can be automatically learned from the unlabeled drug data for representation. \par

If we just initialize the corresponding latent factor in the two-tower model with the drug representation obtained after the above self-supervised learning, this is essentially a pre-training approach that cannot effectively improve the accuracy of the main task (drug-disease association prediction). Therefore, to ensure that the auxiliary task can improve the accuracy of the main task, we leverage a multi-task training strategy where the main task (supervised) and the auxiliary task (self-supervised) are jointly optimized. To be precise, the embedding layer employed in the auxiliary task shares parameters with the drug embedding layer of the two-tower model in the main task. In Section 2.3, we explore why the joint training strategy can contribute to the prediction performance of the main task. It essentially adds a regularization term to the loss function of the main task, which enhances the generalization performance of the main task. \par

In addition, some works use autoencoders as embedding layers for mining the latent factor of drugs and diseases. With the deepening of the network layers in the autoencoder, the problem of information loss occurs, and the decoder cannot restore the original input. However, the idea of the autoencoder is that a good latent factor must be able to restore the original input. Therefore, in order to solve the problem that the latent factor cannot be restored to the original input due to information loss, we design a multi-input decoder. Unlike the previous decoder layer which only accepts the output of the previous network layer, we input the latent factor into each decoder layer to enhance the reconstructing ability of the decoder. Subsequent experiments verify that the latent factor has better prediction performance after the above operation. \par

Our main contributions in this work are as follows. \par

\begin{enumerate}
	\item We present a multi-task self-supervised learning framework for tackling the label sparsity problem in computational drug repositioning. The framework jointly optimizes the main task ((drug-disease association prediction)) and auxiliary task (self-supervised learning) through parameter sharing to improve the representation of drugs, which can be used to improve the prediction performance of potential drug-disease associations.
	\item We improve the decoder in the autoencoder so that each decoder can take into account latent factor information. Thus, a good latent factor can be obtained without loss of information.
	\item On three real-world drug-disease association datasets, we demonstrate that self-supervised learning as an auxiliary task improves the performance of the main model. In addition, it is proved that autoencoders with multi-input decoder have better prediction performance.
\end{enumerate}

The rest of this paper is organized as follows. Section 2 introduces the dataset used in this work and the proposed multi-task self-supervised learning framework and multi-input decoder. Section 3 is the experimental part, including multiple ablation experiments, comparison experiments with state-of-the-art and case study. Section 4 summarizes the work of this paper and discusses future directions of work. \par

\section{Materials and Methods}

In this work, we design a multi-task self-supervised learning framework for solving the problem of label sparsity in computational drug repositioning by learning better drug representations. Firstly, we introduce the dataset used in this work in Section 2.1. Secondly, we introduce the working principle of the two-tower model in computational drug repositioning in Section 2.2. Subsequently, we describe how the multi-task self-supervised learning framework improves the accuracy  of drug-disease prediction in Section 2.3. Finally, in Section 2.4, we describe the workflow of the multi-input decoder.\par

\subsection{Datasets}

In this work, we use three popular real-world datasets \cite{n2}. The drugs in these three datasets are marketed drugs from DrugBank database \cite{drugbank}, diseases from Online Mendelian Inheritance in Man database \cite{omim} and the sparsity of each dataset is different. Table 1 lists the statistics of the above dataset, in which the Gottlieb dataset contains 593 drugs, 313 diseases and 1933 treatment relationships. The Cdataset contains 663 drugs, 409 diseases and 2532 treatment relationships. The DNdataset contains 1490 drugs, 4516 diseases and 1008 treatment relationships. \par

The above dataset also contains similarity matrix between drugs, $DrugSim$, which is calculated from the SMILES chemical structures of the drugs \cite{smile,cdk}. The similarity matrix between diseases is calculated from the medical description information between them. \par

\subsection{The two-tower model in computational drug repositioning}

Computational drug repositioning can be defined as a binary classification problem, given a target drug $i$ and a target disease $j$, we input their respective features into the model $M$, resulting in a prediction value of 0 or 1, where 0 means the drug $i$ is not able to treat the disease $j$ and 1 means the drug $i$ is able to treat the disease $j$. \par

The current popular model of computational drug repositioning is the two-tower model. This framework is using two different sets of neural networks to learn the latent factor of drugs and diseases. Its architecture is shown in Figure 1. In this architecture, the features of the drug and the disease are separately input to an embedding layer containing a series of neural networks for extracting the respective latent factor. The latent factor of the drug and the disease are subsequently computed by a similarity algorithm (inner product) to derive the prediction value $\hat{R}_{ij}$, which represents the probability that the drug can treat the disease. \par

\begin{figure*}[t]
	\centering
	\subfigure[The general architecture of the auxiliary task.]{
		\includegraphics[scale=0.9]{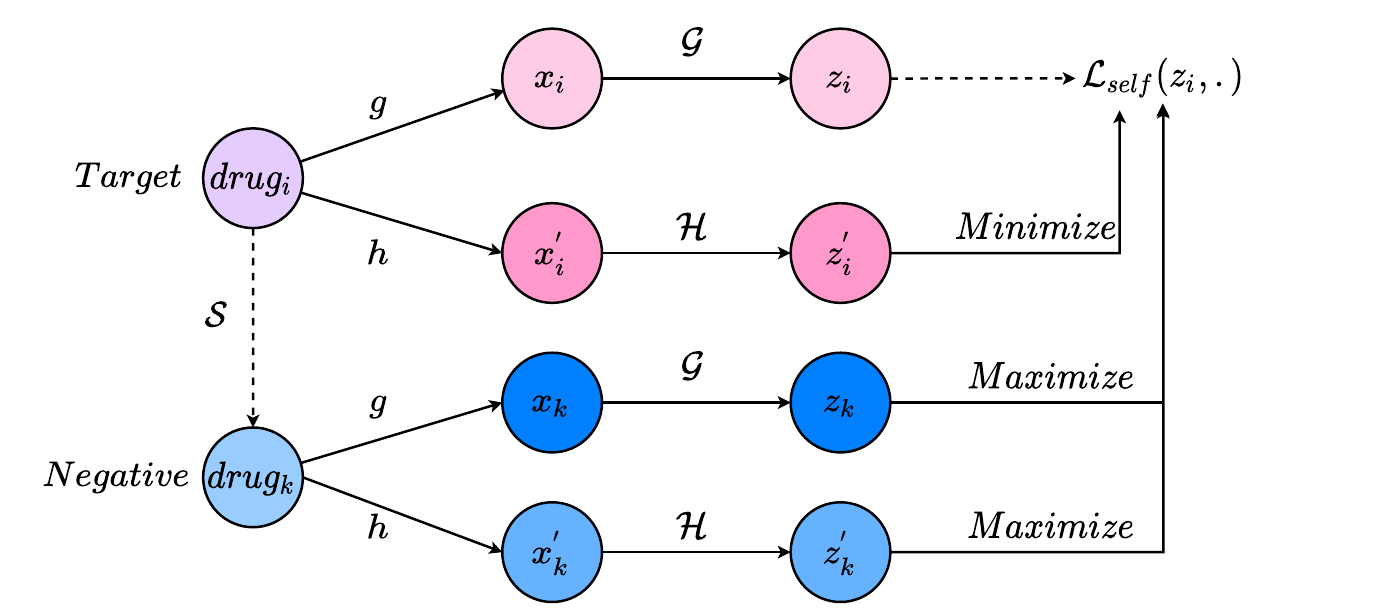}
		%\caption{fig1}
	}
	\quad
	\subfigure[The implementation detail of auxiliary task in this work.]{
		\includegraphics[scale=0.7]{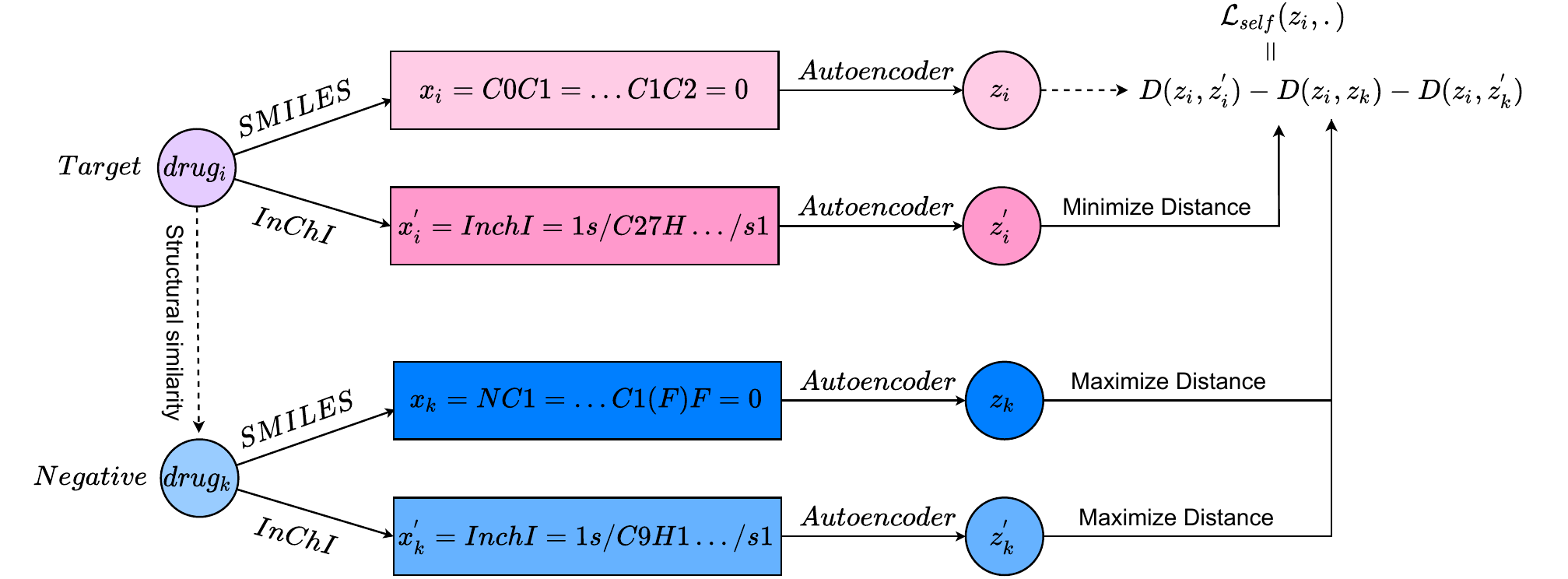}
	}
	
	\centering
	\caption{ The architecture of the auxiliary task.}
\end{figure*}\par

\begin{figure*}[t]
	\centering
	\includegraphics[scale=0.66]{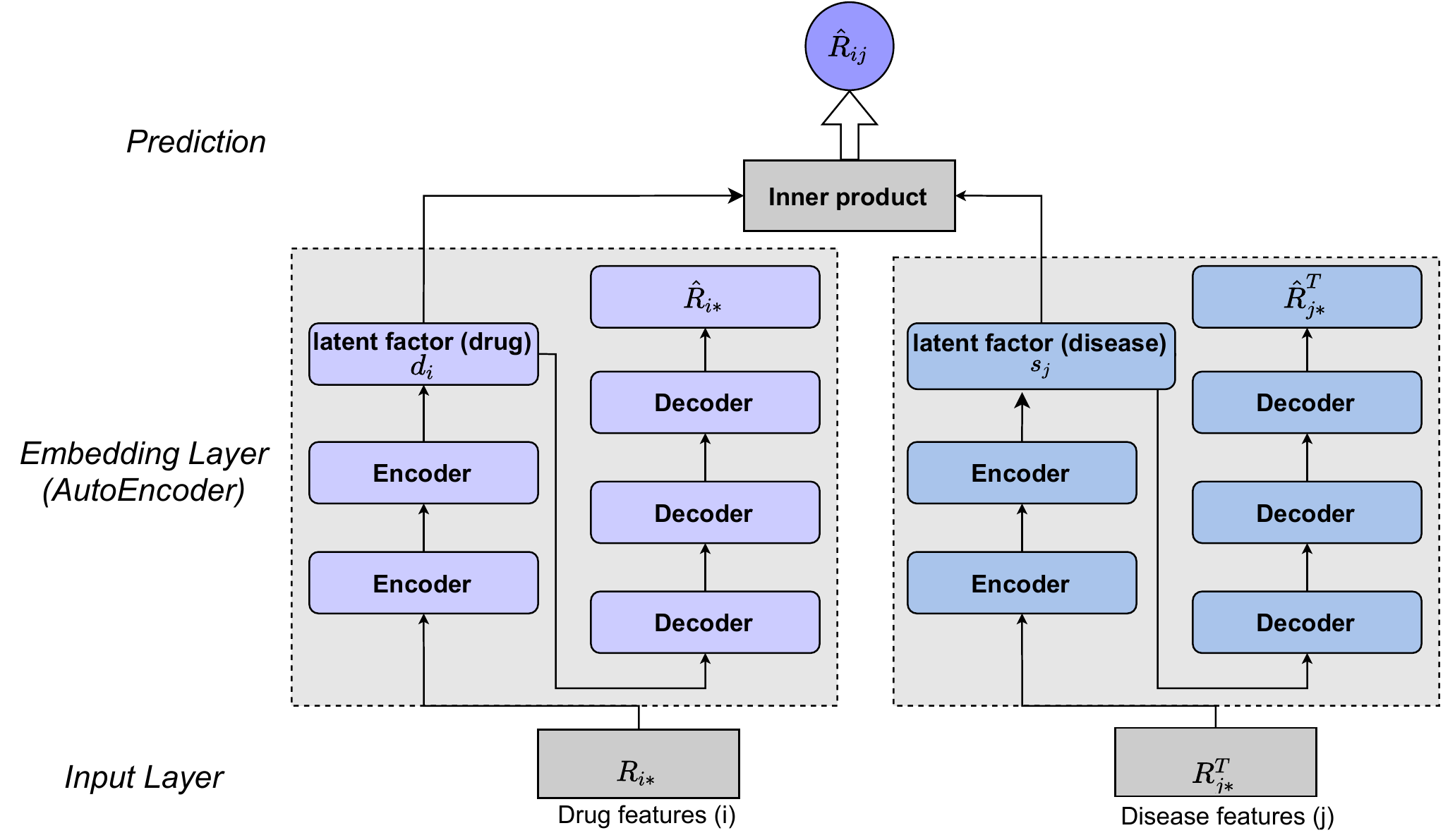}
	\caption{The workflow of main task in the SSLDR framework.}
	\label{}
\end{figure*} 

\subsection{The multi-task self-supervised learning framework}

We present a multi-task self-supervised learning framework for computing drug repositioning. The framework is used to tackle label sparsity by learning a better drug representation. Specifically, we treat the prediction of drug-disease associations as the main task (supervised learning). And automatically mining the internal relationships of drug features is the auxiliary task (self-supervised learning), which aims to learn a good drug representation in the presence of unlabeled data. In this subsection, we first introduce the auxiliary task, and then describe how the auxiliary task improves the accuracy of the main task through a joint training strategy. \par

The auxiliary task of the SSLDR framework can be divided into three steps. The first step is to use the similarity information to filter out the negative neighbors of the target drug. The second step is representation learning for the target drug and negative neighbors using two different data augmentation strategies. The third step is to map the above four drug representations into the embedding space through the embedding layer so that the latent factors belonging to the same drug are close to each other in the embedding space, and those not belonging to the same drug are far away. In this way, a better latent factor can be automatically learned from the unlabeled drug data for representation. Figure 2 (a) shows the general architecture of the auxiliary tasks, and Figure 2 (b) shows the implementation detail of this architecture in this work. \par

Firstly, as shown in equation (1), given a target drug $i$, we select the drug $k$ with the lowest similarity to the target drug $i$ as the negative drug based on the similarity function $S$. In particular, we use the SMILES string of the drug to calculate the similarity information of the target drug to all drugs in the dataset. The similarity information can be obtained from the similarity matrix of drugs $DrugSim$, which can be downloaded from public websites. 
\par

\begin{equation}
	k \leftarrow S(i)
\end{equation}

Subsequently, as shown in equation (2), we do data augmentation for the target drug $i$ and negative drug $k$ using two different transfer functions $h$ and $g$. For the target drug $i$, we want to learn two different representations $x_i$ and $x_i^{'}$ after different data augmentation to ensure that the model still recognizes that $x_i$ and $x_i^{'}$ represent the same drug $i$. Same for the negative drug $k$, two different representations $x_k$ and $x_k^{'}$ are also learned by different data augmentation. \par

\begin{align}
	x_i \leftarrow g(drug_i), \quad x_i^{'} \leftarrow h(drug_i) \\ \notag
	x_k \leftarrow g(drug_k), \quad x_k^{'} \leftarrow h(drug_k)
\end{align}

In practice, we use two different chemical structure representations of drugs as data augmentation strategies. The transfer functions $h$ and $g$ are replaced by the SMILES string of the drug and the International Chemical Identifier (InChI), respectively. Both SMILES strings and InChI represent the chemical structure information of a drug with a small number of characters. And to enable the strings to be input by the deep learning model, we use the Word2Vec algorithm to convert the strings into numeric vectors that can be received by the system. \par

As shown in equation (3), we then input $x_i$ and $x_i^{'}$ into the embedding functions $H$ and $G$, respectively, to obtain $z_i$ and $z_i^{'}$ as the two latent factors of the target drug $i$. The same operation is performed for $x_k$ and $x_k^{'}$ to obtain two latent factors of negative drug $k$, $z_k$ and $z_k^{'}$. In practice, we use the autoencoder model as the embedding function.

\begin{align}
	z_i = \mathcal G(x_i), z_i^{‘} = \mathcal  H(x_i^{‘}) \\ \notag
	z_k = \mathcal G(x_k), z_k^{‘} = \mathcal  H(z_i^{‘})
\end{align}

After obtaining the latent factor of the target drug $i$ and the negative drug $k$, we want to make the distance between $z_i$ and $z_i^{'}$ belonging to the same drug in the embedding space as close as possible. The distance between $z_i$ and $z_k$,$z_k^{'}$ that do not belong to the same drug becomes as far as possible in the embedding space. Therefore, as shown in equation (4), we define the following loss function to let these latent factors contrast themselves. \par

\begin{equation}
	\mathcal L_{auxiliary} = D(z_i,z_i^{'})-D(z_i,z_k) -D(z_i,z_k^{'}) 
\end{equation}

Where $D$ is the distance metric function. In this work, we use the Euclidean distance as the distance metric function. For example, $D(z_i,z_i^{'}) = \| z_i - z_i^{'} \|^2$. \par

\begin{figure*}[t]
	\centering
	\includegraphics[scale=0.54]{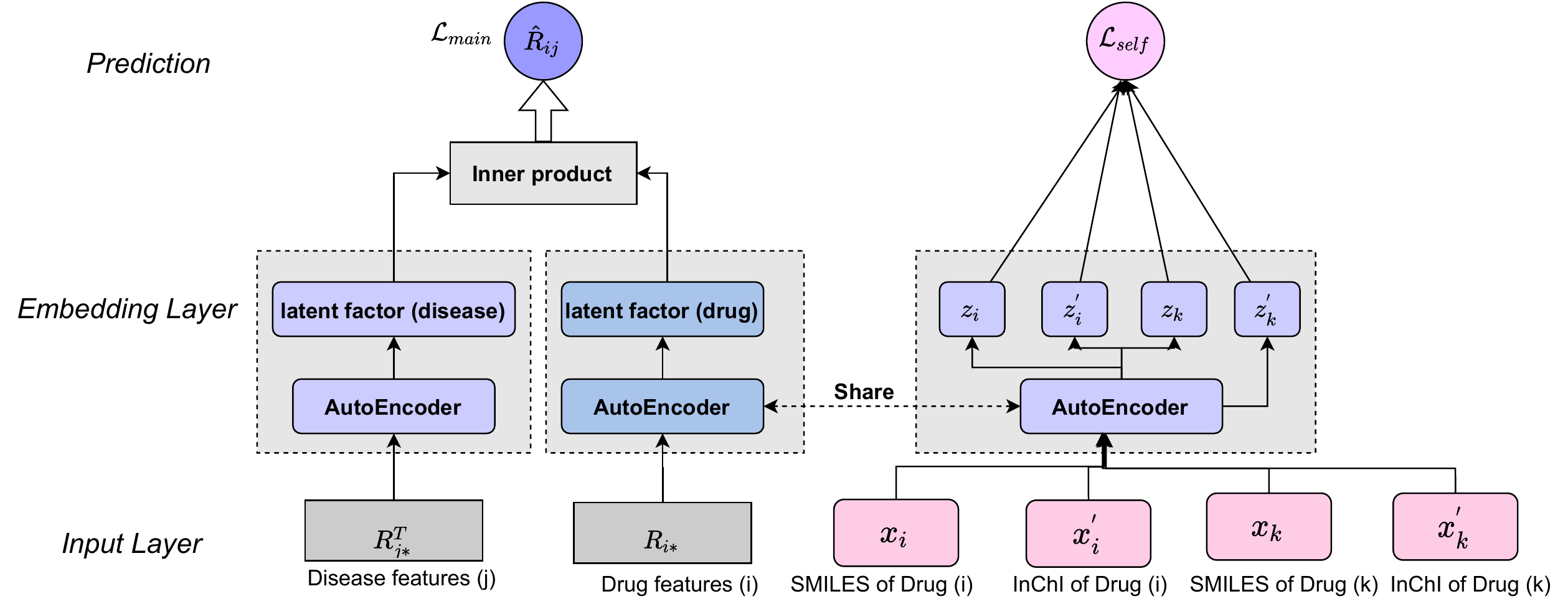}
	\caption{The workflow of multi-task training strategy.}
	\label{}
\end{figure*} 

The main task in the SSLDR framework is based on the two-tower model with the flow chart shown in Figure 3, which contains three modules, namely the input layer, embedding layer and prediction layer. \par

\textbf{Input Layer.}  We take the treatment information of the target drug $i$ for all diseases as its input feature, which is the ith row of $R$, $R_{i*}$. $R$ is the drug-disease association matrix. The treated information of the target disease $j$ for all drugs is used as its input feature, which is the jth column of $R$, $R_{*j}$. The benefit of this feature is the ability to keep a record of behavioral preferences for drugs or diseases.\par

\textbf{Embedding Layer.}   We use a autoencoder model with 2 encoders and 3 decoders as the component to extract the latent factor of a drug (disease). Taking the drug as an example, the operation formula is shown below. \par

\begin{align}
	d_i &= f (W_2^T f(W_1^T R_{i*} + b_1)  + b_2) \\
	\hat{R}_{i*} &= f(V^T_3 f (V_2^T f(V_1^T d_i + b_3)  + b_4) + b_5) \\
	\mathcal L_{drug} &= \|  \hat{R}_{i*} - R_{i*}  \|^2
\end{align}

Where equation (5) is the encoding part, equation (6) is the decoding part, and equation (7) is the loss function of the autoencoder. The $d_i$ is obtained by minimizing $ \mathcal L_{drug} $. Similarly, the latent factor of the disease $j$, $s_j$, can be obtained by minimizing $\mathcal L_{disease}$.\par

\begin{align}
	\mathcal L_{disease} = \|  \hat{R}_{*j} - R_{*j}  \|^2
\end{align}

\textbf{Prediction Layer.} The predicted probability $\hat{R}_{ij}$ that the target drug $i$ can treat the target disease $j$ is obtained by performing the inner product operation on $d_i$ and $s_j$. \par

\begin{align}
	\hat{R}_{ij} = d_i^T s_j
\end{align}

\begin{figure}[h]
	\centering
	\includegraphics[scale=0.7]{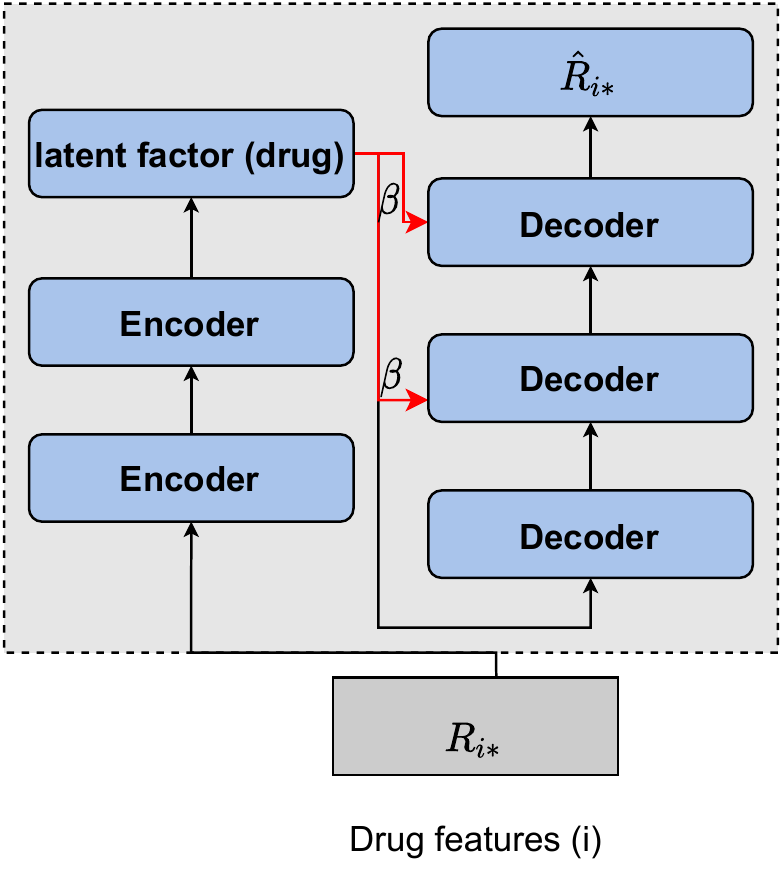}
	\caption{The autoencoders with multi-input decoder.}
	\label{}
\end{figure} 

However, if the latent factor of the drugs learned by the auxiliary task is used as the initial value in the corresponding embedding layer of the main task, this is essentially a pre-training approach that cannot effectively improve the prediction accuracy of the main task. Therefore, to ensure that the drug representation learned by the auxiliary task can improve the prediction accuracy of the main task (drug-disease association prediction), we leverage a multi-task training strategy where the drug-disease association prediction task (supervised) and the auxiliary task (self-supervised) are jointly optimized (Figure 4). To be precise, the embedding layer used by the auxiliary task shares parameters with the embedding layer of the main task, i.e., the same autoencoder model is used. The overall loss function is as follows. \par

\begin{align}
	\mathcal L = \mathcal L_{main} + \alpha \mathcal L_{auxiliary}
\end{align}

Where the loss function $\mathcal L_{main} $ for the main task is shown below. \par
\begin{align}
	\mathcal L_{main} =-\big[ R_{ij} \log \hat{R}_{ij} + (1-R_{ij}) \log (1-\hat{R}_{ij}) \big] +  \mathcal L_{drug} + \mathcal L_{disease}
\end{align}
Where $R_{ij}$ is the label value, $\mathcal L_{drug} $ is the loss when the autoencoder extracts the latent factor of the drug, $\mathcal L_{disease} $ is the loss when the autoencoder extracts the latent factor of the disease. \par

It can be explained from the following perspectives why the joint training strategy can improve the generalization ability of the model. In the form of parameter sharing, it essentially adds a regularization term to the loss function of the main task, which improves the variation range of parameters in the embedding layer, thereby enhancing the generalization performance of the main task. \par

\subsection{The autoencoder with multiple-input decoder}

The two-tower model mentioned above uses the autoencoder to extract latent factors of drugs or diseases. The deeper the number of layers in the autoencoder  model, the more efficient latent factor it can capture. However, according to the logic of the autoencoder model, a good latent factor depends on whether it can restore the original features. After the original features undergo multi-layer encoding and decoding operations, the problem of information loss occurs due to the disappearance of gradients and other reasons. As a result, the latent factor cannot restore the original features. \par

We believe that the biggest reason why the latent factor cannot restore the original input features is that the decoder only receives a single input from the previous layer, which may become sparse as the number of layers deepens. Therefore, based on the decoding architecture of the main task, we additionally add the latent factor to the input of each decoder (Figure 5), so that each layer of the decoder can take into account the information from the latent factor. Its decoding operation is shown in the following equation (12). \par

\begin{align}
	\hat{R}_{i*}= f(V^T_3 ( f (V_2^T ( f(V_1^T d_i + b_3)  + \beta  d_i)+ b_4) + \beta d_i )+ b_5)
\end{align}

Observing the above equation, it can be concluded that when the adjustment parameter $\beta$ is 0, it is the same as the original decoder. However, when the value of the adjustment parameter $\beta$ is greater than 0, the decoder can take into account the information of latent factor and overcome the problem of information loss.\par

\section{Experiments and Discussion}

We provide empirical results on three real-world drug-disease association datasets to demonstrate the effectiveness of our proposed multi-task self-supervised learning framework and multi-input decoder. The experiments designed in this section were used to answer the following research questions. \par

\begin{itemize}
	\item[\textbf{RQ1}] Can our proposed strategy of jointly optimizing the auxiliary task and the main task improve the prediction accuracy of the latter?
	\item[\textbf{RQ2}] Can our proposed autoencoder with multi-input decoder have an advantage in prediction performance compared to traditional autoencoder?
	\item[\textbf{RQ3}] Can our proposed multi-task self-supervised learning framework outperform the state-of-the-art model?
	\item[\textbf{RQ4}] How can our proposed multi-task self-supervised learning framework help in practical applications?
\end{itemize}

\subsection{Evaluation Metrics}

The experiments in this section use 10-fold cross-validation to assess the generalization ability of the model. We first treat the known drug-disease associations as positive samples and divide them into 10 parts equally. We take turns to use 9 of them sequentially as the training set and the remaining 1 as the test set. In addition, we added all unknown drug-disease associations as negative samples to the test set. The parameters in the model are subsequently trained by the training set and the generalization performance of the model is evaluated by the test set. Finally, the average of the computed results of the 10 rounds is calculated, and the value represents the result of the 10-fold cross-validation of the model. \par
Computational drug repositioning is a binary classification problem. In order to fairly compare the generalization performance of the models, we use three popular evaluation metrics for evaluating the performance of the models. They are AUC (Area Under Curve), AUPR (Area Under Precision-Recall Curve) and F1-Score, among which the latter two can more objectively evaluate the performance of the model in the case of positive and negative sample disproportion. \par

\subsection{Parameter Setting}

The values of all hyperparameters in the SSLDR model are chosen based on their performance on the validation set. The validation set is formed by sampling ten percent of the data from the training set. The variation interval of the latent factor vector dimension of drugs and diseases is $[8, 16, 32, 64, 128]$. The variation interval of the parameter $alpha$ in formula (10) is $[0.1, 0.3, 0.5, 0.7, 0.9]$. The variation interval of the parameters of the loss function of the autoencoder is $[0.1, 0.3, 0.5, 0.7, 0.9]$. The learning rate of the model optimizer varies in the interval $[0.1, 0.05, 0.01, 0.005, 0.001]$. In the experiments in this section, the default values for the above parameters are 64, 0.5, 0.5 and 0.001. \par

\begin{table*}[t]
	\centering
	\caption{The experimental results of SSLDR model and SSLDR-M model on three datasets}
	\setlength{\tabcolsep}{2mm}
	\begin{tabular}{@{}c|ccc|ccc|ccc@{}}
		\toprule
		Dataset & \multicolumn{3}{c|}{Gottlieb} & \multicolumn{3}{c|}{Cdataset} & \multicolumn{3}{c}{DNdataset} \\ \midrule
		& AUC     & AUPR    & F1-Score  & AUC     & AUPR    & F1-Score  & AUC     & AUPR    & F1-Score  \\ \midrule
		SSLDR-M & 0.964   & 0.394   & 0.171     & 0.967   & 0.402   & 0.178     & 0.957   & 0.227   & 0.147     \\
		SSLDR   & 0.982   & 0.422   & 0.243     & 0.987   & 0.437   & 0.279     & 0.978   & 0.289   & 0.216     \\ \bottomrule
	\end{tabular}
\end{table*}

\begin{table*}[t]
	\centering
	\caption{The experimental results of SSLDR model and SSLDR-A model on three datasets}
	\setlength{\tabcolsep}{2mm}
	\begin{tabular}{@{}c|ccc|ccc|ccc@{}}
		\toprule
		Dataset & \multicolumn{3}{c|}{Gottlieb} & \multicolumn{3}{c|}{Cdataset} & \multicolumn{3}{c}{DNdataset} \\ \midrule
		& AUC     & AUPR    & F1-Score  & AUC     & AUPR    & F1-Score  & AUC     & AUPR    & F1-Score  \\ \midrule
		SSLDR-A & 0.971   & 0.406   & 0.192     & 0.965   & 0.398   & 0.235     & 0.961   & 0.273   & 0.208     \\
		SSLDR   & 0.982   & 0.422   & 0.243     & 0.987   & 0.437   & 0.279     & 0.978   & 0.289   & 0.216     \\ \bottomrule
	\end{tabular}
\end{table*}

\subsection{Effectiveness of joint optimization of auxiliary task and main task (RQ1)}

\begin{table*}[t]
	\centering
	\caption{The experimental results of the SSLDR model with all the comparison algorithms}
	\setlength{\tabcolsep}{2mm}
	\begin{tabular}{@{}c|ccc|ccc|ccc@{}}
		\toprule
		Dataset & \multicolumn{3}{c|}{Gottlieb} & \multicolumn{3}{c|}{Cdataset} & \multicolumn{3}{c}{DNdataset} \\ \midrule
		& AUC     & AUPR    & F1-Score  & AUC     & AUPR    & F1-Score  & AUC     & AUPR    & F1-Score  \\ \midrule
		MF      & 0.705   & 0.01    & 0.018     & 0.757   & 0.039   & 0.025     & 0.701   & 0.009   & 0.003     \\
		SVM     & 0.724   & 0.007   & 0.021     & 0.763   & 0.044   & 0.027     & 0.752   & 0.007   & 0.036     \\
		NCF     & 0.817   & 0.014   & 0.036     & 0.868   & 0.051   & 0.047     & 0.721   & 0.01    & 0.044     \\
		MLP     & 0.822   & 0.023   & 0.042     & 0.872   & 0.052   & 0.038     & 0.736   & 0.011   & 0.051     \\
		ASMF    & 0.891   & 0.165   & 0.221     & 0.911   & 0.232   & 0.264     & 0.856   & 0.051   & 0.107     \\
		NMFDR   & 0.887   & 0.27    & 0.232     & 0.927   & 0.267   & 0.278     & 0.821   & 0.057   & 0.112     \\
		SSLDR   & 0.982   & 0.422   & 0.243     & 0.987   & 0.437   & 0.279     & 0.978   & 0.289   & 0.216     \\ \bottomrule
	\end{tabular}
\end{table*}

To answer RQ1, we evaluate the impact of the auxiliary tasks on the main task under the joint training strategy. We compare SSLDR with the following baseline. The baseline is SSLDR-M, a variant of the SSLDR model with auxiliary tasks removed. By comparing the experimental results of SSLDR and SSLDR-M, we can intuitively compare whether the auxiliary task can improve the prediction accuracy of the main task. \par

Table 2 shows the experimental results of the SSLDR model and the SSLDR-M model on the three real-world datasets. Firstly, it can be intuitively found from the table 2 that the SSLDR model outperforms the SSLDR-M model on all metrics and datasets. On the three datasets, the average improvement under the AUC, AUPR and F1-Score metrics are 2\%, 14.3\% and 48.5\%, especially the improvement under the F1-Score metric is the most obvious. The above results illustrate that the loss function of the auxiliary task is used as the regularization term of the loss function of the main task by a joint training strategy, which optimizes the search space of the parameters in the main task. This enables the model of the main task to have better generalization performance. \par

Furthermore, we find that the performance gap between the SSLDR model and the SSLDR-M model is proportional to the sparsity between the datasets, i.e., the greater the sparsity of the datasets, the greater the performance gap between the SSLDR model and the SSLDR-M model. The sparsity of Gottlieb dataset, Cdataset and DNdataset increases in turn, and compared with the SSLDR-M model, the average improvement of SSLDR on these three datasets is 17\%, 22.5\% and 25.4\%, respectively. This is because the parameters in the SSLDR-M model rely on labeled data for training. The small amount of labeled data in the sparse dataset prevents SSLDR-M model learning effective latent factor of drug and disease. And the SSLDR model additionally uses self-supervision and joint optimization to ensure that the main task learns a better latent factor of the drug. Therefore, the main task can have a better prediction effect. The discussion of the above experimental results proves that the auxiliary task can improve the prediction accuracy of the main task, indicating the correctness of our improvement points. \par

\subsection{Effectiveness of autoencoder with multi-input decoder (RQ2)}

To answer RQ2, we evaluate the prediction performance of the autoencoder with multi-input decoder. We compare SSLDR with the following baseline. The baseline is SSLDR-A, a variant of SSLDR, which uses the original autoencoder to replace the autoencoder with multiple-input decoder in the SSLDR model. Specifically, the decoder of the original autoencoder only receives input from the previous network layer. Through the direct comparison between SSLDR and SSLDR-A, we can verify whether the multi-input decoder can overcome the problem of information loss, thereby improving the prediction ability of the latent factor. \par

Table 3 shows the experimental results of SSLDR model and SSLDR-A model on three datasets. It can be significantly found that the SSLDR model with a multi-input decoder outperforms the SSLDR-A model with the single-input decoder on all metrics and datasets. On the three datasets, the average improvements under the AUC, AUPR and F1-Score metrics are 1.7\%, 6.5\% and 16.3\%. The above experimental results show that adding the latent factor to the input of each decoder allows it to take into account the information from the latent factor, so that the model can learn a better latent factor, thereby improving its expressiveness and the predictive power for drug-disease associations. \par

\begin{table*}[t]
	\centering
	\caption{List of diseases recommended by SSLDR}
	\setlength{\tabcolsep}{1mm}
	\begin{tabular}{@{}c|c@{}}
		\toprule
		Drugs (DrugBank IDs)                   & Top 5 candidate diseases (OMIM IDs)                                  \\ \midrule
		\multirow{5}{*}{Doxorubicin (DB00997)} & \textbf{Kaposi's sarcoma, susceptibility (148000)}; \\
		& Cutaneous Malignant, Susceptibility to, 1 (155600);                  \\
		& MELANOMA, CUTANEOUS MALIGNANT, SUSCEPTIBILITY TO, 2; CMM2 (155601)   \\
		& Testicular Germ Cell Tumor (273300);                                 \\
		& \textbf{Prostate Cancer (176807)};                  \\ \midrule
		\multirow{5}{*}{Gemcitabine (DB00441)} & \textbf{Prostate Cancer (176807)};                  \\
		& Mismatch Repair Cancer Syndrome (276300);                            \\
		& Hereditary Diffuse (137215);                                         \\
		& UTERINE ANOMALIES (192000);                                          \\
		& \textbf{Colorectal Cancer (114500) };               \\ \midrule
		\multirow{5}{*}{Vincristine (DB00541)} & Testicular Germ Cell Tumor (273300);                                 \\
		& \textbf{Small Cell Cancer of The Lung (182280)};    \\
		& \textbf{Kaposi's sarcoma, susceptibility (148000)}; \\
		& \textbf{Breast Cancer (114480)};                    \\
		& Bladder Cancer (109800);                                             \\ \bottomrule
	\end{tabular}
\end{table*}

\subsection{Comparison of experimental results (RQ3)}

To answer RQ3, we compare the experimental results of the SSLDR model with the following mainstream computational drug repositioning models. \par

\begin{itemize}

	\item \emph{MF} \cite{mf}: The matrix factorization model uses the inner product and the latent factor to infer the probability of a therapeutic relationship between the drug and the disease.
	\item \emph{SVM} \cite{svm}: The support vector machine (SVM) are currently popular binary classification models.
	\item \emph{NCF} \cite{ncf}: The Neural Collaborative Filtering uses neural networks and Hadamard products to uncover potential new uses for drugs.
	\item \emph{MLP} \cite{mlp}: The multi-layer perceptron (MLP) consists of multiple neural networks and sigmoid activation functions for binary classification problems.
	\item \emph{ASMF} \cite{m3}: The ASMF model utilizes an attention mechanism instead of the inner product, enabling the model to take into account the unique weights of each feature. And they use the improved autoencoder model to extract more effective features of drugs and diseases.
	\item \emph{NMFDR} \cite{m4}: The NMFDR model used a modified Euclidean distance and point space to represent the drug-disease relationship.
	
\end{itemize}

Table 4 presents the experimental results of the SSLDR model with all the comparison algorithms. The following conclusions can be drawn from observing the table. \par
The MF and NCF, as primary latent fator-based models, achieved poor prediction results because they were not able to incorporate more auxiliary information about drugs or diseases. And the ASMF model incorporates drug or disease similarity information, it substantially outperforms the MF and NCF model models on all datasets and metrics. And on the three datasets, the SSLDR model outperforms SVM (machine learning model) and MLP (deep learning model). \par

It can be clearly found that the prediction results of the SSLDR model are better than the two SOTA models, ASMF and NMFDR, in all datasets and all indicators. Compared with the NMFDR model, the average improvement of SSLDR under the AUC metric on the three datasets is 12.1\%, the average improvement under the AUPR metric is 175\%, and the average improvement under the F1-Score metric is 32.6\%.  \par

It is not difficult to find that through the contrastive learning between the two enhanced representations of the drug, the model can learn a better drug representation, thereby improving the generalization ability of the model. In addition, the additional input of the decoding layer enhances the reconstruction ability of the model. This also enhances the representation of drugs and diseases to a certain extent, thereby improving the predictive power of the model. \par

\subsection{Case study (RQ4)}

We selected 3 drugs from the Gottlieb dataset to validate the usefulness of SSLDR in practical applications.  These three drugs are doxorubicin, gemcitabine and vincristine, all of which are used to treat oncology diseases. Oncological diseases are currently the focus of research and development by pharmaceutical companies, so it is of great value to find potential therapeutic drugs for these diseases. \par

Table 5 lists the diseases recommended by the SSLDR model for these three drugs. The bolded diseases in the table indicate that they have been verified in the CTD dataset to have a therapeutic relationship with the corresponding drugs. For the drugs doxorubicin and gemcitabine, two new diseases were correctly predicted, and both were hits in the first and fifth spots. The last drug, vincristine, has 3 diseases that are correctly recommended in the list of recommended diseases. \par

The results of the above case studies show that compared with the previous computational drug repositioning models, the disease list recommended by SSLDR model has a higher hit rate, and most diseases are successfully predicted under the condition of higher ranking. Therefore, this can significantly accelerate the process of drug screening and research and development, and has great economic and practical value for practical application scenarios. \par

\section{Conclusion}
In this work, we propose a multi-task self-supervised learning framework SSLDR for the problem of label sparseness in computational drug repositioning. Under the strategy of joint training, the framework uses auxiliary tasks to improve the latent factor of drugs to enhance the generalization performance of the main task "drug-disease association prediction". And we propose a multi-input decoder to improve the ability of the autoencoder to mine latent factors of drugs or diseases. Experimental results on multiple real-world datasets validate the superiority of the multi-task self-supervised learning framework and multi-input decoder. \par

For future work, we plan to explore how to improve the latent factor of disease so that it can be better applied to computational drug repositioning scenarios. In addition, the framework proposed in this work is based on the matrix factorization model, and how to apply this framework to graph-based model is also the direction of our future work. \par

% \usepackage{multirow}

%begin{thebibliography}{99}  

%bibitem{1} 

%end{thebibliography}
%\cite{*}
\bibliography{reference.bib}

\end{document}